\documentclass[letterpaper]{article} % DO NOT CHANGE THIS
\usepackage{aaai2027}
\nocopyright

\usepackage[hyphens]{url}  % DO NOT CHANGE THIS
\usepackage{graphicx} % DO NOT CHANGE THIS
\usepackage{natbib}  % DO NOT CHANGE THIS AND DO NOT ADD OPTIONS
\usepackage{caption} % DO NOT CHANGE THIS AND DO NOT ADD OPTIONS
\usepackage{algorithm}
\usepackage{algorithmic}
\usepackage{newfloat}
\usepackage{listings}
\DeclareCaptionStyle{ruled}{labelfont=normalfont,labelsep=colon,strut=off}
\floatstyle{ruled}
\newfloat{listing}{tb}{lst}{}
\floatname{listing}{Listing}
\usepackage{booktabs}
\usepackage{amsmath}
\usepackage{amssymb}

\title{EnergyBridge: Benchmarking Household Energy Management, User Participation, and Grid Flexibility}
\author{
Xudong Wu\equalcontrib\textsuperscript{\rm 1},
Zeqing Wu\equalcontrib\textsuperscript{\rm 2},
Jiarui Zhang\textsuperscript{\rm 2},
Xuhao Fan\textsuperscript{\rm 1},
Ziang Ding\textsuperscript{\rm 1},\\
Yuming Zhuang\textsuperscript{\rm 3},
Mingqi Yuan\textsuperscript{\rm 1},
Yilun Du\textsuperscript{\rm 2},
Hongjie Jia\corresponding\textsuperscript{\rm 2},
Yunfei Mu\corresponding\textsuperscript{\rm 2},
Jiayu Chen\corresponding\textsuperscript{\rm 1}
}

\affiliations{
\textsuperscript{\rm 1}The University of Hong Kong\\
\textsuperscript{\rm 2}Tianjin University\\
\textsuperscript{\rm 3}Beihang University
}

\begin{document}
\maketitle
\begin{abstract}

Residential virtual power plants (VPPs) can provide grid flexibility by
shifting household demand, but physical flexibility becomes dependable capacity
only when residents authorize a plan and the promised response is delivered.
Existing benchmarks evaluate control but omit event-specific authorization. We
present EnergyBridge, a benchmark and agent framework connecting capacity
reporting, household authorization,  and physical execution. It combines region-specific EnergyPlus environments for
Tianjin, and 
Berlin with an LLM-based User Participation Simulator. Against 584
persona- and event-matched human role-play judgments, the LLM-based User Participation Simulator preserves
method ordering with a 5.3-point mean absolute acceptance error. Across
conventional controllers and agent baselines, EnergyBridge achieves the highest
simulated authorization, lowest event-window energy, and the most reliable capacity commitment in both regions. We release human data and codes for reproducible human-centered grid-flexibility
research: \url{https://github.com/Agentic-Intelligence-Lab/EnergyBridge}.
\end{abstract}

% ============================================================
% 1. Introduction
% ============================================================
\section{Introduction}
\label{sec:introduction}Variable renewable generation, growing electrified loads, and local demand
peaks are increasing the need for short-duration grid flexibility. Demand
response enables households to provide such flexibility by temporarily
reducing, shifting, or rescheduling flexible loads, including heating,
ventilation, and air conditioning (HVAC), electric vehicles (EVs), water
heaters, and task-oriented appliances \cite{Martinez2022,Sajjad2016}. By
aggregating distributed household responses, virtual power plants (VPPs)
create grid-scale flexibility portfolios that can provide
demand-side capacity when needed \cite{Wang2020,nationalgrideso2021crowdflex}.
However, converting household flexibility into reliable grid resources requires
more than physical capability: households must authorize the requested actions,
and the committed capacity must be delivered in practice.

Household participation is inherently uncertain because decisions depend on
comfort preferences, domestic routines, expected incentives, and tolerance for
external automation. Field studies show substantial variation in residential
response across different events and automation settings
\cite{nationalgrideso2021crowdflex}. Therefore,
households should be modeled as active decision makers rather than passive
flexibility assets. A VPP must determine not only whether a physical action is
possible, but also whether a household will accept and sustain that action.

Beyond authorization, a VPP must estimate how much flexibility each household
can reliably provide before an event. Existing capacity estimation approaches
typically rely on device states, historical demand, weather conditions, event
timing, and aggregation across households \cite{Sajjad2016,Wang2020}. While
aggregation can reduce random errors, it cannot eliminate systematic
uncertainties caused by shared weather conditions, behavioral patterns, or
comfort constraints. For
example, an air conditioner may have theoretical load reduction potential, but
that capacity is not reliable if occupants reject the corresponding comfort
change. Thus, accurate flexibility reporting requires household and event
context rather than aggregation alone.

A key question is therefore how to model household participation preferences: given a flexibility request and its expected consequences, will a household accept, revise, or reject the proposed action, and how can this participation decision inform reliable flexibility delivery? Persona-conditioned large language model (LLM) simulators provide a scalable approach for representing such user decisions. Prior studies show that role-conditioned LLMs can
reproduce aggregate human response patterns in controlled experiments
\cite{Aher2023,Argyle2023SiliconSamples}, while matched human evaluations can
further calibrate task-specific reliability
\cite{Liu2026TravelPersona,Dou2025SimulatorArena}. However, existing VPP
benchmarks mainly focus on physical flexibility, whereas agent-based HEMS
studies emphasize interaction and preference reasoning. They lack a unified
evaluation framework that connects household authorization, personalized
capacity reporting, execution decisions, and measured grid delivery
\cite{ElMakroum2026,Jung2026,Dolinin2026SmartHouseOperator}.

\begin{figure*}[ht]
    \centering
    \includegraphics[width=1\textwidth]{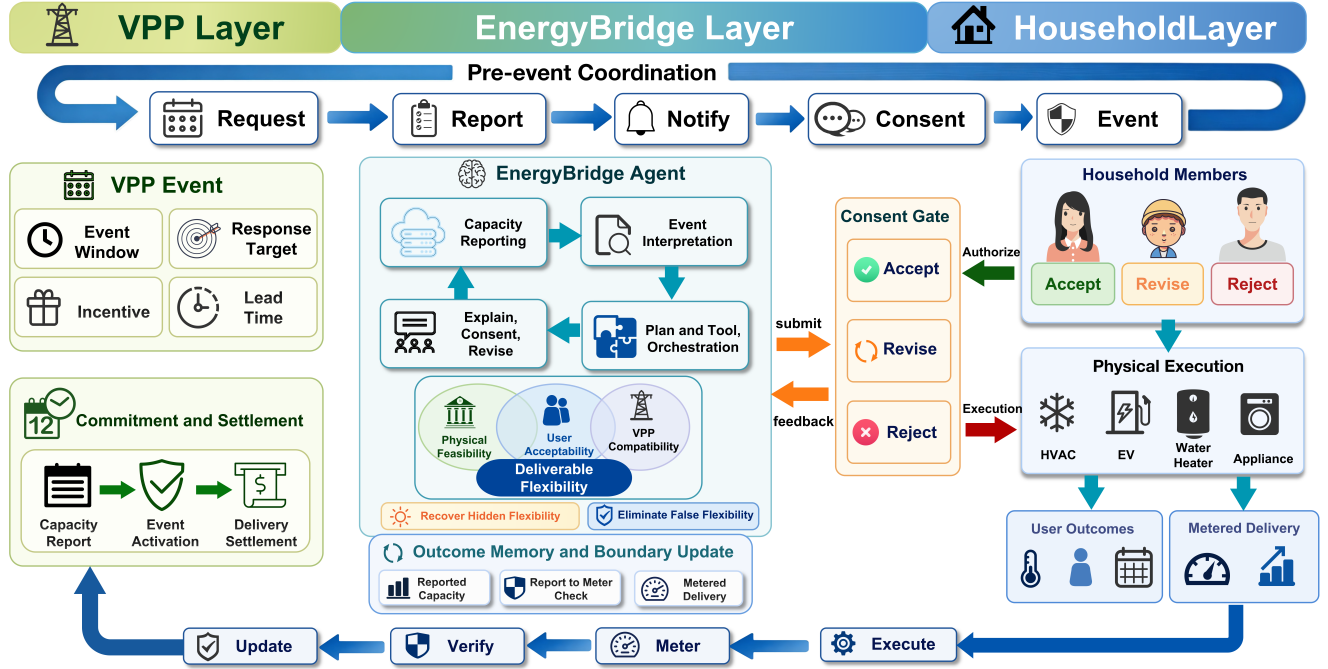}
    \caption{Overview of the EnergyBridge benchmark and agent.
Before the event, the VPP sends a flexibility request to the household. The EnergyBridge agent first combines household preferences, device states, and outcomes from previous events to generate an pre-event capacity report, and then invokes control tools to produce a candidate plan and a household-facing explanation. Household members may accept, revise, or reject the plan. An accepted plan is executed; revision feedback is stored in memory and used to regenerate the proposal; and rejection triggers the ordinary fallback policy. During execution, EnergyPlus simulates HVAC, EV charging, water heating, and household appliances. The benchmark then compares the pre-event reported capacity with the realized delivered capacity and feeds the verification results back into the agent’s memory. This workflow enables EnergyBridge to evaluate the complete chain from VPP requests and household authorization to physical execution and capacity delivery within a unified framework.}
    \label{fig:task_overview}
\end{figure*}

To bridge this gap, we introduce EnergyBridge, a benchmark and agent framework
that connects household decisions with VPP-level flexibility delivery.
EnergyBridge operates at the household--VPP interface, where a household-level
agent translates VPP flexibility requests into household-level decisions
and actions.

EnergyBridge evaluates the complete process. Given a VPP
flexibility request, the agent estimates and reports pre-event
capacity, generates a feasible device-level plan under operational constraints,
and requests resident authorization. The approved plan is executed in an EnergyPlus simulation environment, while revision triggers plan regeneration and
rejection activates the fallback policy. The realized delivery is then compared
with the pre-event capacity report. This design separates physical capability,
user-authorized flexibility, and actual delivered flexibility under a common
evaluation protocol.

The agent uses questionnaires, calendars, device states, historical feedback,
and event memory to learn household preferences and personalize both control
decisions and capacity reports. Instead of treating households as fixed
resources, it continuously updates its understanding of what each household can
accept and how much flexibility the VPP can reliably offer.

We evaluate EnergyBridge using regional building simulations, human-grounded
authorization evaluation, and post-event evaluation of capacity-reporting
accuracy. The benchmark compares EnergyBridge with conventional controllers
and existing agent-based methods under the same household context, execution
protocol, and metrics. Results show that EnergyBridge improves household
participation while achieving reliable VPP-side flexibility reporting,
demonstrating the importance of connecting human authorization with physical
energy delivery.

Our contributions are:
\begin{itemize}
    \item \textbf{An end-to-end benchmark for human-authorized physical agents.}
EnergyBridge formulates residential flexibility as a closed-loop agent task:
an agent must reason over household preferences, plan under device and
operational constraints, request household authorization, execute only
the authorized plan in a physical simulator, and ensure that its pre-event
capacity report matches the realized delivery. This enables joint evaluation
of user interaction, constrained decision making, physical execution, and
outcome reliability, rather than evaluating language responses or control
performance in isolation.

    \item \textbf{Human-grounded evaluation and open resources.}
    EnergyBridge incorporates human-calibrated user simulation and releases
    benchmark environments, human evaluation data, physical models, and
    baseline implementations to support reproducible research.

    \item \textbf{A cross-layer flexibility agent.}
    The EnergyBridge agent combines household grounding, constraint-aware plan
    generation, consent-gated execution, and personalized capacity reporting,
    enabling reliable conversion from household flexibility potential to
    delivered grid resources.
\end{itemize}

\section{Related Work}
\label{sec:related_work}

\subsection{Benchmarks for Energy Management}
Existing building and community energy benchmarks primarily focus on
standardizing physical control, economic evaluation, and grid-interactive
operation. BOPTEST provides reproducible building-control tests, and CityLearn
evaluates flexibility, resilience-oriented control, and carbon-aware community
energy management
\cite{Blum2021,Nweye2025}. These environments mainly characterize the physical outcomes after control
execution without modeling household decisions. This omission is consequential
in flexibility settings, because a technically feasible control
action may still be revised or rejected by occupants, directly affecting
whether the promised flexibility is delivered.

Modeling user decision therefore requires a complementary evaluation
mechanism. Persona-conditioned language models have shown the ability to reproduce
heterogeneous user preferences and choices in controlled settings
\cite{Aher2023,Argyle2023SiliconSamples}. Large-scale simulation studies further
show that LLM-based representative samples can approximate aggregate human
responses in social science experiments
\cite{Ashokkumar2026SocialScience}. Task-specific reliability can then be
calibrated and evaluated through persona- and context-matched human judgments
\cite{Liu2026TravelPersona,Dou2025SimulatorArena}.

EnergyBridge builds on these findings by introducing a benchmark-owned User
Participation Simulator and calibrating its simulated decisions with matched
human response data.

\subsection{Conventional and Language-Model HEMS Control}

MPC, MILP, and reinforcement-learning based HEMS optimize energy cost,
comfort, and appliance operation under predefined objectives and constraints
\cite{Han2023,Amer2023,Mu2023RollingOptimization,ElMakroum2026LoadScheduling}.
These methods provide strong physical modeling and reliable device-level
control. However, household preferences are typically represented through
manually designed parameters rather than learned from historical interactions
and contextual household information. Therefore, conventional HEMS lacks the
ability to continuously adapt to heterogeneous user preferences and provide
personalized flexibility decisions.

Recent language-model and multi-agent HEMS extend energy management with
natural-language interaction and personalized reasoning. Agentic AI HEMS
translates household preferences, day-ahead prices, and calendar constraints
into appliance schedules \cite{ElMakroum2026}. HEMA and SmartHouseOperator
further explore multi-agent assistance, user confirmation, memory, and device
interaction \cite{Jung2026,Dolinin2026SmartHouseOperator}. These systems
demonstrate the potential of LLM-based agents for personalized and interactive
energy management. However, their quantitative evaluations mainly focus on the quality of agent-generated strategies, with limited validation through closed-loop
execution in high-fidelity, physics-based building simulation environments.

EnergyBridge Agent combines preference-aware decision making with
physics-grounded execution, enabling personalized and reliable flexibility
control.

% ============================================================
% 2. EnergyBridge
% ============================================================
\section{EnergyBridge Benchmark}
\label{sec:energybridge}

Hereafter, \emph{EnergyBridge benchmark} refers to the common evaluation
framework, while \emph{EnergyBridge agent} refers to our proposed method.
As shown in Figure~\ref{fig:task_overview}, the benchmark covers the full process
from a VPP request and household consent to physical execution and delivery
verification. This section defines the request-to-delivery task, introduces the physical
household environment and user participation simulator, and specifies the
benchmark outputs and metrics.

\subsection{Task Definition}
\label{sec:task}

Residential VPP flexibility requires more than physical controllability.
A household device may provide technically feasible flexibility, but the
corresponding action may not be authorized by residents or successfully
delivered after execution. EnergyBridge therefore evaluates agents through a
complete request-to-delivery process connecting capacity reporting, household
authorization, physical execution, and delivery verification.

Given a VPP flexibility request, an evaluated method must:
(1) estimate the capacity that can be reliably offered before execution,
(2) generate a household-feasible flexibility plan,
(3) obtain household authorization through the consent gate,
(4) execute the confirmed plan or ordinary fallback branch, and
(5) verify the realized flexibility from physical trajectories.

This protocol distinguishes three levels of flexibility capacity.Let
$F^{\mathrm{phys}}_{i,e}$ denote the physically feasible flexibility of
household $i$ during event $e$ under comfort, safety, and service constraints.
The aggregate physical flexibility is

\begin{equation}
C^{\mathrm{phys}}_e
=
\sum_{i\in I_e}F^{\mathrm{phys}}_{i,e}.
\end{equation}

After considering household decisions, only a fraction of this flexibility
is authorized:

\begin{equation}
C^{\mathrm{auth}}_e
=
\sum_{i\in I_e}
\alpha_{i,e}F^{\mathrm{phys}}_{i,e},
\end{equation}

where $\alpha_{i,e}\in[0,1]$ represents the event-specific authorization
fraction. After execution, the baseline-adjusted metered trajectory determines
the delivered flexibility:

\begin{equation}
C^{\mathrm{del}}_e
=
\sum_{i\in I_e}
\eta_{i,e}\alpha_{i,e}F^{\mathrm{phys}}_{i,e},
\end{equation}

where $\eta_{i,e}$ captures the realization gap between authorized and
delivered flexibility.

Therefore, EnergyBridge evaluates whether an agent can transform physical
flexibility potential into household-authorized and verifiable grid resources,
rather than only generating feasible control schedules.

\subsection{Physical Household Simulator}

\paragraph{Building dynamics.}
EnergyBridge uses EnergyPlus 25.1, a whole-building energy simulation engine
\cite{Crawley2001}. Tianjin and 
Berlin instantiate separate residential
building models whose envelope parameters follow GB~50176-2016 and
DIN~4108-4:2020-11, respectively
\cite{MOHURD2016GB50176,DIN2020DIN4108}. Each model is simulated at six zone
timesteps per hour (10 minutes) and exposes zone temperature, outdoor
conditions, HVAC actuation, facility electricity, and event-window meter data.

\paragraph{Appliances.}
The household layer models HVAC, EV charging, an electric water heater, washer,
dryer, and dishwasher, together with non-controllable base load. Its interface
permits household-specific thermal-zone use, appliance availability, and device
settings. EV service is constrained by arrival, departure, charging power, and
target state of charge. The water heater uses a shiftable-load service proxy
with an availability window and nominal setpoint. Task appliances have duration,
precedence where applicable, and completion deadlines. At every EnergyPlus
timestep, the appliance simulator writes electrical loads back to the physical
run.

\paragraph{Regional context.}
Tianjin uses a normalized time-of-use tariff and the CSWD typical-year EPW for
WMO station 545270 \cite{CMA2005CSWD}.\footnote{Tianjin CSWD listing:
\url{https://climate.onebuilding.org/WMO_Region_2_Asia/CHN_China/index.html};
file \texttt{CHN\_TJ\_Tianjin.545270\_CSWD.zip}.}
Berlin uses hourly 2025 station observations from the Deutscher Wetterdienst
Climate Data Center \cite{DWD2026HourlyClimate} and hourly 2025 price traces. The
main physical runs use occupancy schedules to determine device availability and
service constraints. 

\subsection{User Participation Simulator}

\paragraph{Household representation.}
Each base persona is represented along six behavior dimensions:
\begin{equation}
p=(p^{\mathrm{schedule}},p^{\mathrm{comfort}},p^{\mathrm{task}},
p^{\mathrm{price}},p^{\mathrm{control}},p^{\mathrm{grid}}).
\end{equation}
These dimensions control routines, comfort tolerance, task flexibility, price
sensitivity, automation trust, confirmation requirements, and willingness to
support the grid. The main benchmark composes them into five multi-member
households (Appendix Table~\ref{tab:households}). Shared devices are scheduled once,
while members retain distinct calendars and evaluation criteria.

Prior studies show that persona-conditioned LLMs can reproduce aggregate human
response patterns in controlled settings
\cite{Aher2023,Argyle2023SiliconSamples}. Profile-conditioned role-play models
can also align their ratings with human judgments \cite{Dou2025SimulatorArena}.
Following empirical persona-alignment approaches \cite{Liu2026TravelPersona},
we designed the six-dimensional role cards and evaluator prompt, refined them
with formative participant feedback, and froze both before validation. The
prompt exposes temperature, routines, service deadlines, explanation
specificity, and override-relevant conflicts established in residential
demand-response studies \cite{Sarran2021,Kaspar2024,Nambiar2025}.

Section~\ref{sec:human_grounding} directly evaluates aggregate authorization
alignment against 584 human responses under matched roles, events, question
wording, and method-name-masked consequence payloads. It recovers the same
method ordering with a 5.3-percentage-point acceptance MAE
(Table~\ref{tab:human_calibration}).

\paragraph{Pre-execution participation.}
Before execution, household members are not told which method generated the
candidate plan. They are shown only the proposed device schedule and a summary
of its expected consequences. For each member, the Role-Play Household
Simulator produces an acceptance probability, a response label
(accept, conditional, or reject), and structured feedback. Conditional
responses specify constraints for revising the proposal. After revision, the
members' acceptance probabilities are averaged to obtain the household-level
acceptance probability. A fixed household-event random draw then converts this
probability into a reproducible accept-or-reject decision.

\paragraph{Post-event scoring.}
After execution, the identity of the method remains hidden from household
members. They are shown only the realized outcomes, including indoor
temperature, device schedules, task completion, VPP conflicts, and the
user-facing explanation. Each member then assigns a score, and the household
score is computed as the arithmetic mean across all member scores.

\subsection{Benchmark Outputs and Metrics}
\label{sec:benchmark_metrics}

The benchmark reports three groups of method-independent outcomes.
\begin{itemize}
    \item \textbf{User outcomes:} simulated gate-acceptance rate
    $A=N_{\mathrm{accepted}}/N_{\mathrm{events}}$, mean post-event household
    score on a 1-5 scale.
    \item \textbf{Household physical and service outcomes:} thermal outcomes,
    task completion, daily energy cost, and VPP-window energy.
    \item \textbf{Capacity and safety outcomes:} event-authorized capacity and delivered capacity.
\end{itemize}

% ============================================================
% 3. EnergyBridge Agent
% ============================================================
\section{EnergyBridge Agent}
\label{sec:agent}

This section describes the three main components of the EnergyBridge agent:
household memory, personalized plan synthesis, and retrieval-based capacity
reporting.

\subsection{Household Knowledge and Memory}

EnergyBridge uses a two-level hierarchical memory to limit the amount of
information included in the planning context.
The first level stores a structured household profile initialized from a
four-question onboarding questionnaire. The questionnaire responses are
normalized into preferences and planning rules covering comfort, task
flexibility, trust in automation, calendar constraints, and price or grid
priorities. The second level stores episodic information from previous events,
including proposed plans, authorization decisions, executed actions, delivery
outcomes, evaluation scores, and user feedback.

Before each event, the planner receives the household profile, the current
planning rules, a compact summary of prior interactions, and the three most
recent event records. After execution, new feedback and outcomes are used to
update the preference rules and the compact summary for subsequent decisions.

\subsection{Plan Synthesis}

EnergyBridge treats conventional control methods as callable skills, including
Model Predictive Control (MPC) and rule-guided Mixed-Integer Linear Programming
(MILP). Each skill returns a normalized candidate record containing device
actions. The agent selects and personalizes candidate plans using household memory and historical feedback. The physical validator rejects unsafe action records.

The resulting plan is presented in household-relevant terms. 
Using household memory and historical feedback, the agent generates
personalized explanations and revision suggestions to address individual
concerns and improve user acceptance.

\subsection{Retrieval-Augmented Capacity Reporting}

EnergyBridge retrieves the $k$ most similar historical events using region,
household type, event time and duration, pre-event load, weather, available
devices, and schedule context. Historical delivery is baseline-adjusted to the
target event. Only records preceding the target period are eligible. Their
adjusted deliveries form an empirical distribution from which the pre-event
report is selected before the consent decision. We refer to this
estimator as retrieval-based capacity reporting. Reporting  conditional
reporting accuracy and reliable capacity coverage defined in
Section~\ref{sec:capacity}.

\begin{figure*}[!t]
    \centering
    \includegraphics[width=\textwidth]{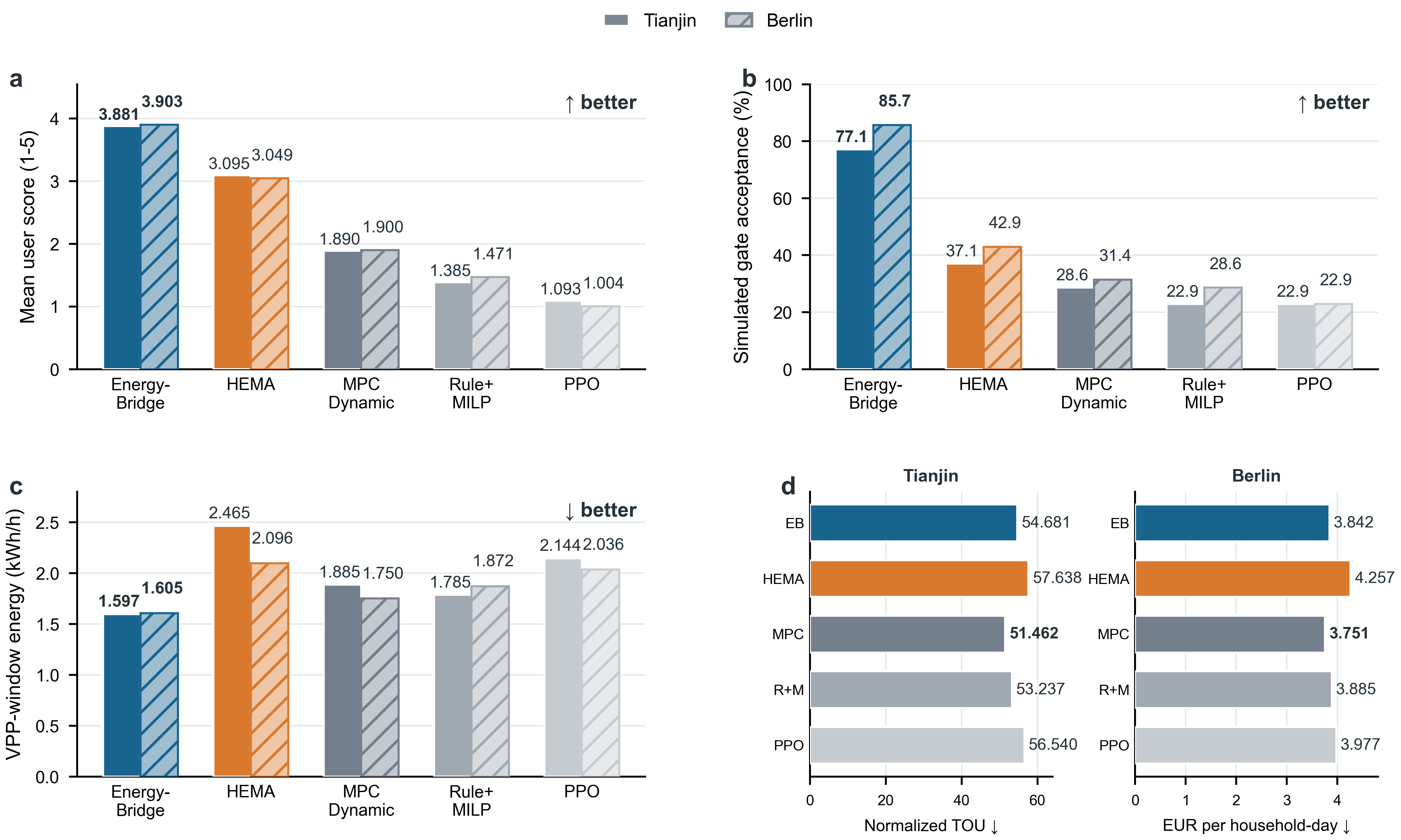}
    \caption{Region-specific results ($n=35$ events per region-method). Score and
    gate acceptance are simulated. Tianjin cost uses normalized TOU units,
    Berlin cost is EUR per household-day, and VPP energy is kWh per event hour.}
    \label{tab:main_results}
\end{figure*}

% ============================================================
% 4. Experimental Evaluation
% ============================================================
\section{Experimental Evaluation}
\label{sec:experimental_setting}

We evaluate EnergyBridge at three levels. First, we examine whether the
agent can achieve desirable household outcomes. Second,
we compare simulated user authorization with 584 matched human responses to
validate the reliability of the participation simulator. Third, we evaluate
the accuracy of pre-event flexibility reports.

\subsection{Baselines}

\paragraph{HEMA agent.}
HEMA is an open multi-agent home-energy assistant with analysis, knowledge, and
control tools \cite{Jung2026}. We retain its design, including native routing, tool graph, memory,
natural-language feedback, and event explanations, making it the closest native
agent comparison.

\paragraph{Physical-control references.}
The three controller baselines cover established method families: building-energy
MPC \cite{Mu2023RollingOptimization,Slaifstein2026,Behzadi2026}, household MILP
scheduling \cite{deLima2026,ElMakroum2026LoadScheduling}, and DRL-based building
control \cite{Wu2022FutureInformation,Wu2026AETD3}.
MPC Dynamic optimizes energy, comfort, appliance constraints, and event-window
load over a one-hour horizon. Rule+MILP protects fixed services with interpretable
rules and schedules flexible services by mixed-integer optimization. We
implement PPO \cite{Schulman2017} for DRL-based
 building control. Detailed formulations of the three controller baselines are provided in Appendix~\ref{app:baseline_details}.

\subsection{Main Evaluation}

The main matrix crosses five households, two regions, seven days, and five
methods. One 18:00-19:00 event per household-day yields 70 matched contexts per
method and 350 episodes. Tianjin and 
Berlin use distinct regional building
models. Within each matched household-region-day context, all methods share the same household configuration and external conditions, with differences arising only from the evaluated method.

Their interaction budgets and tool usage are recorded under their original configurations, with token consumption comparisons reported in Table~\ref{tab:llm_resources}.

EnergyBridge achieves the highest simulated user score and
acceptance rate, the lowest VPP-window energy, and 100\% modeled task completion
(Figure~\ref{tab:main_results}). Relative to HEMA, the simulated acceptance gap is
40.0 percentage points in Tianjin and 42.9 points in  Berlin, while event-window
energy is 35.2\% and 23.4\% lower, respectively.

\subsection{Personalization Capability}
\begin{table*}[!t]
\centering
\setlength{\tabcolsep}{3pt}

\begin{tabular}{lrrrr@{\hspace{1em}}lrrrr}
\toprule
\multicolumn{5}{c}{Price-sensitive} &
\multicolumn{5}{c}{Comfort-sensitive} \\
\cmidrule(lr){1-5}
\cmidrule(lr){6-10}

Method
& Score $\uparrow$
& Cost $\downarrow$
& $^\circ$C
& kWh $\downarrow$
&
Method
& Score $\uparrow$
& Cost $\downarrow$
& $^\circ$C
& kWh $\downarrow$ \\

\midrule

EnergyBridge
& 4.21 & 23.431 & 27.5 & 1.224
&
EnergyBridge
& 4.51 & 32.252 & 25.5 & 1.634 \\

HEMA
& 4.00 & 32.551 & 23.9 & 1.344
&
HEMA
& 2.37 & 32.965 & 23.9 & 3.840 \\

MPC Dynamic
& 4.21 & 26.454 & 27.0 & 1.437
&
MPC Dynamic
& 2.57 & 26.454 & 27.0 & 1.437 \\

Rule+MILP
& 4.40 & 15.590 & 28.0 & 1.113
&
Rule+MILP
& 2.10 & 17.701 & 28.0 & 1.774 \\

PPO
& 3.22 & 20.314 & 27.1 & 1.263
&
PPO
& 2.57 & 20.810 & 27.0 & 1.283 \\

\bottomrule
\end{tabular}

\caption{Controlled persona evaluation with acceptance and fallback disabled.
Cost is normalized Tianjin TOU; VPP kWh is integrated over the one-hour event.
Higher score is better; preferred setpoint is persona-dependent.}
\label{tab:persona_adaptation}
\end{table*}

We evaluate whether agents can adapt executable control decisions to
different household preferences and routines. Table~\ref{tab:persona_adaptation}
shows that EnergyBridge provides the strongest personalization capability
among the evaluated methods. Additional calendar cases in
Appendix Table~\ref{tab:calendar_adaptation} show that EnergyBridge further
adapts schedules according to occupancy, return-home time, care requirements,
and service deadlines. These results demonstrate that household context
changes the executable control policy rather than only the generated
explanation in EnergyBridge.

\subsection{Contribution of Household Grounding Components}

To identify which components enable EnergyBridge's high authorization rate,
we ablate the household grounding information used for personalized proposals
and consent decisions. The fixed-plan replay keeps the event, physical plan,
household, and random draw unchanged while removing one component at a time.
With the complete EnergyBridge context, simulated acceptance reaches 77.1\%.
Removing feedback memory reduces acceptance to 65.7\%, removing calendar
context reduces it to 54.3\%, removing the questionnaire-based household
profile reduces it to 45.7\%, and replacing personalized explanations with a
generic explanation reduces it to 37.1\% .
These results show that household grounding and memory-based personalization
directly affect authorization decisions, while physical feasibility remains
unchanged.

\subsection{Human Grounding of Role-Play Authorization}
\label{sec:human_grounding}
This experiment tests whether human and LLM role players authorize the same
methods under matched persona and event information. Human respondents adopt
an assigned household persona and evaluate the same frozen role card, Tianjin
event, question wording, and method-name-masked consequence trajectory shown
to the LLM evaluator. The comparison therefore isolates role-play
authorization rather than respondents' own household preferences or method
identity. Formative feedback refined the six role cards and evaluator prompt
before both were frozen for validation.

The cohort contains 584 complete consented responses across the six personas
($n=97,116,73,88,91,119$). Each respondent evaluates MPC, HEMA, and
EnergyBridge, yielding 1,752 paired respondent-method judgments. The primary
endpoint is method-level acceptance because it determines entry into
consent-gated execution. Following aggregate human-LLM choice alignment
\cite{Liu2026TravelPersona}, we evaluate aggregate rather than individual
prediction. We report Wilson intervals for human rates, paired respondent
bootstrap intervals for method contrasts, and Bernoulli JSD and acceptance MAE
for human-LLM alignment.

The cohort contains 293 female and 291 male respondents, with ages spanning
20--55 years. Across persona forms, female shares range from 46.4\% to 54.8\%,
and mean ages range from 29.5 to 34.3 years.
Coarse source geolocation covers 21 province-level regions and 80 city labels
(Appendix Table~\ref{tab:human_demographics}).

\begin{table}[!t]
\centering
\setlength{\tabcolsep}{3pt}

\begin{tabular}{lrrrr}
\toprule
Method & LLM$\uparrow$ & Human$\uparrow$ [95\% CI] &
Gap$\downarrow$ & JSD$\downarrow$ \\
\midrule
MPC & 28.6\% & 25.7\% [22.3, 29.4] & 2.9 & 0.0005 \\
HEMA & 37.1\% & 45.2\% [41.2, 49.3] & 8.1 & 0.0034 \\
EnergyBridge & 77.1\% & 72.1\% [68.3, 75.6] & 5.1 & 0.0017 \\
\bottomrule
\end{tabular}

\caption{Prompt- and event-matched aggregate human--LLM authorization
alignment. Human rates use 584 responses; LLM rates use 35 frozen Tianjin
evaluator records per method. Brackets are Wilson 95\% intervals. Gap is the
absolute difference between unrounded rates in percentage points.}
\label{tab:human_calibration}
\end{table}

Human and LLM aggregate judgments produce the same acceptance ordering:
EnergyBridge, HEMA, then MPC (Table~\ref{tab:human_calibration}). The EnergyBridge advantage over HEMA holds across all six
personas, ranging from 15.5 to 37.4 points
(Appendix Table~\ref{tab:human_persona_alignment}). Excluding responses with a
persona mismatch or completion time below 60 seconds preserves the ordering
(Appendix Table~\ref{tab:human_qc_sensitivity}). These matched human judgments
ground the aggregate authorization ordering used by the role-play benchmark. Further details are provided in Appendix~\ref{Details_of_Human}.

\subsection{Capacity Reporting}
\label{sec:capacity}

Can a capacity value declared before an event be verified after household
authorization selects the physical execution branch? We evaluate this question
in a separate June-to-July capacity audit with 70 held-out events per method.
June delivery records form each method's retrieval memory, while July events
form the query cohort. Every method uses top-5 same-method retrieval. Simulated
consent selects either the proposed controller trajectory or the household's
ordinary fallback before physical execution.

We report three complementary quantities. Conditional reporting
accuracy $B_m$ is the fraction of accepted events whose reported
capacity is verified within the target band. The 20\% tolerance criterion follows the capacity assessment requirement 
 \cite{jiangsu2024demandresponse}. Acceptance rate $A_m$
is the fraction of capacity requests authorized by the simulated
household. We define reliable capacity coverage $F_m$ as the fraction
of all capacity requests that are both authorized and accurately
reported:
\begin{align}
B_m
&=
P\left(
0.8 \leq
\frac{C^{\mathrm{actual}}}{\widehat{C}}
\leq 1.2
\,\middle|\,
\mathrm{accepted}
\right), \\
A_m
&=
P(\mathrm{accepted}), \\
F_m
&=
P\left(
\mathrm{accepted}
\cap
\left\{
0.8 \leq
\frac{C^{\mathrm{actual}}}{\widehat{C}}
\leq 1.2
\right\}
\right).
\end{align}

 An accurate capacity
report does not constitute dispatchable grid capacity if the
household rejects the associated action. Conversely, a highly
accepted request is not reliable if its reported capacity cannot be
physically delivered within the target band.

\begin{table}[t]
\centering
\setlength{\tabcolsep}{3pt}

\begin{tabular}{lrrr}
\toprule
Method & Acc. $B_m$ $\uparrow$ & Accept. $A_m$ $\uparrow$&  $F_m$ $\uparrow$\\
\midrule
EnergyBridge & 90.0\% & \textbf{71.4\%} & \textbf{64.3\%} \\
HEMA & 69.2\% & 37.1\% & 25.7\% \\
MPC Dynamic & 63.2\% & 27.1\% & 17.1\% \\
Rule+MILP & \textbf{93.8\%} & 22.9\% & 21.4\% \\
PPO & 70.6\% & 24.3\% & 17.1\% \\
\bottomrule
\end{tabular}

\caption{Cross-method capacity reporting under method-neutral simulated
consent ($N_m=70$ per method). Overall accurate coverage is calculated as
conditional accuracy multiplied by acceptance rate.}
\label{tab:capacity_heldout}

\end{table}

% ============================================================
% 6. Conclusion
% ============================================================
\section{Conclusion}

Existing residential energy research often evaluates three stages of
flexibility provision separately: pre-event capacity reporting, household
authorization, and physical delivery. However, these stages are sequentially
dependent. Reported capacity is useful to a VPP only if the corresponding
household authorizes the proposed action and the approved plan subsequently
delivers the promised response. Evaluating any stage in isolation therefore
cannot determine whether household flexibility constitutes a reliable grid
resource.

EnergyBridge connects these stages within a complete request-to-delivery
process. Household authorization decisions are simulated by a human-calibrated
User Simulator based on the household context and the proposed
plan. Approved plans
are executed in regional EnergyPlus environments, and the pre-event capacity
report is compared with realized delivery after the event.
The resulting authorization decisions, delivery outcomes, and user feedback
are then recorded and used to update subsequent planning and capacity
reporting.

Across two regional EnergyPlus environments, the EnergyBridge agent achieves
the highest simulated authorization rate and the lowest event-window energy
among the evaluated conventional controllers and agent baselines. The study
with 584 human role-play responses supports the simulated method ordering, and
the post-event audit evaluates the accuracy of capacity reports for accepted
events. These results demonstrate the value of evaluating household
authorization, physical execution, and capacity-reporting accuracy as a single
closed-loop process. We release the benchmark environments, human evaluation
data, physical models, and baseline implementations to support reproducible
research.

\bibliography{aaai2027}

\appendix

\clearpage

\section{Household Persona and Consent Details}
\label{Details_of_Human}
Six base roles span cooperative price-aware commuting, comfort-gated home work,
irregular cautious routines, ideal participation, caregiving/low flexibility,
and EV optimization. Atom personas add comfort sensitivity, automatic-control
trust, price indifference, and task rigidity. The five main households compose
three to five members from these roles.

\begin{table*}[t]
\centering

\begin{tabular}{p{0.23\textwidth}p{0.29\textwidth}p{0.40\textwidth}}
\toprule
Household & Members represented & Principal conflicts tested \\
\midrule
Dual commuter & EV commuter, automation-trusting commuter, student, older adult
& Evening return, study comfort, chores, hot water, and next-day EV readiness.\\
Multigeneration caregiver & Caregiver, older adult, commuter, home worker,
child/student proxy & Vulnerable-member comfort and hot water take priority over
aggressive grid response.\\
Hybrid work from home & Home worker, price-aware commuter, irregular resident
& Daytime occupancy, work comfort, evening EV service, and uncertain schedules.\\
Flexible EV commuter & EV owner, flexible partner, convenience-oriented member
& High physical flexibility subject to a hard mobility deadline.\\
Shared roommates & EV owner, shift worker, price-indifferent member, rigid-task
member & Shared-resource conflicts and partial-veto behavior under irregular
occupancy.\\
\bottomrule
\end{tabular}
\caption{Main household scenarios. All scenarios include HVAC, washer, dryer,
dishwasher, water heater, EV, and base load; the distinction is the member and
routine structure.}
\label{tab:households}
\end{table*}

The human role-play comparison remains consistent within each assigned
persona. EnergyBridge exceeds HEMA in all six paired comparisons, with
persona-specific differences of 15.5--37.4 percentage points
(Table~\ref{tab:human_persona_alignment}).

\begin{table*}[t]
\centering

\begin{tabular}{lrrrrrl}
\toprule
Persona & $n$ & MPC & HEMA & EB & EB--HEMA [95\% CI] & Holm $p$ \\
\midrule
Price-sensitive & 97 & 29.9 & 55.7 & 71.1 & 15.5 [4.1, 26.8] & .0135 \\
Comfort-sensitive & 116 & 24.1 & 44.0 & 68.1 & 24.1 [12.1, 36.2] & $<.001$ \\
Irregular routine & 73 & 19.2 & 43.8 & 63.0 & 19.2 [2.7, 35.6] & .0385 \\
Cooperative regular & 88 & 26.1 & 38.6 & 72.7 & 34.1 [18.2, 48.9] & $<.001$ \\
Caregiver & 91 & 28.6 & 42.9 & 80.2 & 37.4 [24.2, 50.5] & $<.001$ \\
EV commuter & 119 & 25.2 & 45.4 & 75.6 & 30.3 [17.6, 42.0] & $<.001$ \\
\bottomrule
\end{tabular}
\caption{Human authorization by assigned persona. Rates are percentages.
Differences and paired bootstrap intervals compare EnergyBridge (EB) with
HEMA. Reported $p$ values are exact McNemar tests with Holm adjustment across
the three within-persona method contrasts.}
\label{tab:human_persona_alignment}
\end{table*}

\begin{table}[t]
\centering

\resizebox{\columnwidth}{!}{
\begin{tabular}{lrrrr}
\toprule
Subset & $n$ & MPC & HEMA & EnergyBridge \\
\midrule
Full cohort & 584 & 25.7\% & 45.2\% & 72.1\% \\
Persona match & 577 & 25.3\% & 44.5\% & 71.9\% \\
Completion $\geq60$ s & 576 & 24.8\% & 44.4\% & 71.7\% \\
Combined filter & 570 & 24.4\% & 43.9\% & 71.6\% \\
\bottomrule
\end{tabular}}
\caption{Authorization-rate sensitivity to response-quality filters. All
three method decisions and scored outcomes are complete in every subset.}
\label{tab:human_qc_sensitivity}
\end{table}

\begin{table*}[t]
\centering

\begin{tabular}{lrrrrr}
\toprule
Assigned persona & $n$ & Female (\%) & Age, mean $\pm$ SD & Median [IQR] & Regions/cities \\
\midrule
All & 584 & 50.2 & $32.2\pm10.9$ & 26 [23, 41] & 21/80 \\
Price-sensitive & 97 & 46.4 & $32.4\pm11.7$ & 26 [23, 44] & 16/47 \\
Comfort-sensitive & 116 & 50.0 & $30.9\pm9.7$ & 26 [23, 38] & 14/45 \\
Irregular routine & 73 & 54.8 & $34.3\pm11.8$ & 34 [23, 44] & 13/35 \\
Cooperative regular & 88 & 50.0 & $33.9\pm12.0$ & 28 [24, 47.3] & 15/45 \\
Caregiver & 91 & 51.6 & $33.6\pm11.0$ & 30 [24, 43] & 17/47 \\
EV commuter & 119 & 49.6 & $29.5\pm9.0$ & 25 [23, 35] & 15/54 \\
\bottomrule
\end{tabular}
\caption{Distribution of the 584 retained human role players. Gender and age
are self-reported. Region and city counts use coarse source geolocation parsed
before raw IP removal. IQR denotes the interquartile range.}
\label{tab:human_demographics}
\end{table*}

The following calendar cases use a separate controlled set of routine anchors
and therefore complement, rather than duplicate, the two-persona setpoint test
in Table~\ref{tab:persona_adaptation}.

\begin{table*}[t]
\centering

\resizebox{\textwidth}{!}{
\begin{tabular}{p{0.13\textwidth}p{0.18\textwidth}p{0.29\textwidth}rrrrr}
\toprule
Persona & Routine anchors & EnergyBridge realized schedule & EB & HEMA & MPC & R+M & PPO \\
\midrule
Commuter, price-aware & Return 18.5; hot water due 21
& W 19-21; DW 21-22.5; WH 14-18 & 4.18 & 3.82 & 4.21 & \textbf{4.40} & 3.22 \\
Stay-home, comfort & Home during event; hot water due 21
& W 10-12; WH 18-20 & \textbf{3.82} & 2.37 & 2.44 & 2.66 & 2.64 \\
Irregular, cautious & Return 19 $\pm$ 3; hot water due 21
& W 19-21; WH 18-20 & 3.16 & 3.30 & 2.18 & \textbf{3.67} & 3.51 \\
Caregiver & Occupied; vulnerable member; hot water due 20
& W 10-12; WH 17-19 & 3.80 & \textbf{3.82} & 2.42 & 3.71 & 2.51 \\
\bottomrule
\end{tabular}
}
\caption{Calendar adaptation in the one-day no-gate controlled setting. Times
are hours of day; W, DW, and WH denote washer, dishwasher, and water-heater
operation. The five rightmost columns are realized user scores.}
\label{tab:calendar_adaptation}
\end{table*}

\section{Exact Metric Definitions}

For event set $E$, simulated gate acceptance is
\begin{equation}
A=|E|^{-1}\sum_{e\in E}\mathbb{I}[g_e=\mathrm{accept}].
\end{equation}
For household members $M_e$, post-event user score is
\begin{equation}
U_e=|M_e|^{-1}\sum_{j\in M_e}u_{e,j}, \qquad u_{e,j}\in[1,5].
\end{equation}
VPP-window energy is metered from EnergyPlus and appliance loads:
\begin{equation}
E_e^{\mathrm{VPP}}=\sum_{t\in[t_s,t_e)}P_t\Delta t.
\end{equation}
Realized shed uses the event baseline estimator only when valid:
\begin{equation}
C_e^{\mathrm{actual}}=\max(0,E_e^{\mathrm{baseline}}-E_e^{\mathrm{VPP}}).
\end{equation}
For accepted events, this meter deviation is the delivered capacity
$C_e^{\mathrm{del}}$. For rejected events, event-authorized and delivered
capacity are zero, while $C_e^{\mathrm{actual}}$ remains an audit quantity for
the ordinary-fallback trajectory.
Counterfactual no-DR values remain diagnostic and are not silently substituted
for actual shed. Capacity pass $B_e$ is defined in
Section~\ref{sec:benchmark_metrics}.
Task completion is the fraction of required services whose physical completion
and deadline conditions are satisfied. Execution-without-consent counts any
VPP-specific actuation on a rejected branch; fallback restoration checks that
the ordinary fallback is reinstated.

\section{Details of Physical-Control Baselines}
\label{app:baseline_details}

\subsection{Common Physical-Control Interface}

MPC Dynamic, Rule+MILP, and PPO use the same physical execution interface.
At each 10-minute decision step, the controller receives the current simulation
time, indoor and outdoor temperatures, current HVAC setpoint, occupancy proxy,
electricity price, active VPP event and target, appliance configuration,
appliance execution states, and available load or capacity estimates. Each
controller returns a normalized action record containing an HVAC cooling
setpoint and schedules for the controllable household services:
washer, dishwasher, dryer, electric water heater (EWH), and electric vehicle
(EV). Commands for absent devices are discarded by the execution adapter.

\subsection{MPC Dynamic}
\label{sec:mpc_dynamic_details}

\paragraph{State and prediction horizon.}
MPC Dynamic operates at a 10-minute resolution,
$\Delta t=1/6$ hour. Its default prediction horizon is six control steps,
corresponding to one hour:
\begin{equation}
H=6,\qquad H\Delta t=1\ \mathrm{hour}.
\end{equation}
At decision step $k$, the internal state is
\begin{equation}
\begin{split}
s_k=\big(&T_k^{a},T_k^{m},T_k^{e},P_{k-1}^{\mathrm{HVAC,raw}},
z_k^{\mathrm{EV}},p_k^{\mathrm{EV,home}},T_k^{\mathrm{EWH}},\\
& p_{j,k}^{\mathrm{wait}},p_{j,k}^{\mathrm{run}},
p_{j,k}^{\mathrm{finish}},r_{j,k},
p_k^{\mathrm{occ}},\xi_k\big),
\end{split}
\end{equation}
where $T^a$, $T^m$, and $T^e$ are the air, internal-mass, and envelope
temperatures; $z^{\mathrm{EV}}$ is EV state of charge; $T^{\mathrm{EWH}}$
is tank temperature; and the task variables describe the expected
waiting, running, finished, and remaining states of washer, dishwasher, and
dryer. The exogenous vector $\xi_k$ contains outdoor temperature, direct and
diffuse solar radiation, electricity price, base load, time, day type, and VPP
context. At the beginning of each rollout, the unobserved mass and envelope
temperatures are initialized to the measured indoor air temperature. Regional
thermal and HVAC coefficients are selected separately for Tianjin and Berlin.

The prediction model combines a regional 5R3C thermal model, device-state
models, and an expected household-behavior Markov model. The thermal transition
is
\begin{align}
C_a\frac{T^a_{k+1}-T^a_k}{\Delta t}
={}&
g_{oa}(T_k^{o}-T_k^a)
+g_{am}(T_k^m-T_k^a)
\nonumber\\
&+
g_{ae}(T_k^e-T_k^a)
+Q_k^{\mathrm{HVAC,eff}} \nonumber\\
&+r_aQ_k^{\mathrm{sol}},
\label{eq:air_thermal}\\[2pt]
C_m\frac{T^m_{k+1}-T^m_k}{\Delta t}
={}&
g_{am}(T_k^a-T_k^m)
+g_{me}(T_k^e-T_k^m)
\nonumber\\
&+
r_mQ_k^{\mathrm{sol}},
\label{eq:mass_thermal}\\[2pt]
C_e\frac{T^e_{k+1}-T^e_k}{\Delta t}
={}&
g_{ae}(T_k^a-T_k^e)
+g_{me}(T_k^m-T_k^e)
\nonumber\\
&+
g_{eo}(T_k^o-T_k^e)
+r_eQ_k^{\mathrm{sol}}.
\label{eq:envelope_thermal}
\end{align}
The cooling and heating fractions are computed as
\begin{align}
u_k^c&=\operatorname{clip}
 \left(\frac{T_k^a-T_k^{c,\mathrm{sp}}}{2},0,1\right),\\
u_k^h&=\operatorname{clip}
 \left(\frac{T_k^{h,\mathrm{sp}}-T_k^a}{2},0,1\right).
\end{align}
Simultaneous heating and cooling are prohibited by retaining only the larger
fraction. HVAC electrical power is
\begin{equation}
\begin{aligned}
P_k^{\mathrm{HVAC}}
={}&
\frac{\bar Q_c u_k^c}{\mathrm{COP}_c}
+\frac{\bar Q_h u_k^h}{\mathrm{COP}_h}\\
&\quad
+0.25\max\!\left(u_k^c,u_k^h\right).
\end{aligned}
\end{equation}
and the effective thermal input additionally contains the fitted cooling,
heating, fan, and one-step HVAC-lag coefficients.

The EV transition is
\begin{align}
P_k^{\mathrm{EV}}
={}&\min\left\{
\bar P^{\mathrm{EV}}u_k^{\mathrm{EV}}
p_k^{\mathrm{request}}p_k^{\mathrm{home}},
\frac{(z^{\mathrm{target}}-z_k^{\mathrm{EV}})_+C^{\mathrm{EV}}}
{\eta^{\mathrm{EV}}\Delta t}
\right\},\\
z_{k+1}^{\mathrm{EV}}
={}&\operatorname{clip}\left(
z_k^{\mathrm{EV}}+
\frac{\eta^{\mathrm{EV}}P_k^{\mathrm{EV}}\Delta t}
{C^{\mathrm{EV}}},0,z^{\max}\right).
\end{align}
The EWH model accounts for scheduled hot-water draws, standing thermal loss,
rated heating power, thermal efficiency, and minimum and maximum tank
temperatures. For each shiftable task $j$, the expected transition updates the
waiting, running, and finished probabilities. A task can start only after its
earliest start time and when its remaining duration fits before its latest
finish time.

Occupancy, EV-at-home probability, device requests, and preferred HVAC
setpoints are propagated using expected values from the stored factorized
behavior-transition and action-policy matrices. Within a one-hour rollout, the
current outdoor temperature, price, and base-load forecast are retained unless
an explicit forecast is available; solar input follows the implementation's
day/night forecast profile.

\paragraph{Action space and search.}
The action contains
\begin{equation}
\begin{aligned}
a_k=\Bigl(
&T_k^{c,\mathrm{sp}},
\{s_{j,k},b_{j,k}^{\mathrm{skip}}\}_{j},\\
&s_k^{\mathrm{EWH}},e_k^{\mathrm{EWH}},
T_k^{\mathrm{EWH,sp}},\\
&s_k^{\mathrm{EV}},e_k^{\mathrm{EV}}
\Bigr).
\end{aligned}
\end{equation}
The implementation uses sequential discrete enumeration rather than a
continuous nonlinear solver. It first selects the HVAC setpoint, then washer,
dishwasher, and dryer schedules, followed by the EV and EWH schedules. Each
component is retained before the next component is optimized.

\begin{table*}[t]
\centering
\caption{MPC Dynamic action candidates. Only present and currently unlocked
devices are enumerated.}
\label{tab:mpc_action_candidates}
\small
\setlength{\tabcolsep}{4pt}
\renewcommand{\arraystretch}{1.12}
\begin{tabular}{p{0.18\textwidth}p{0.73\textwidth}}
\toprule
Component & Candidate construction \\
\midrule
HVAC &
Candidates include $25.5$, $26.0$, $26.5$, $27.0$, and $27.5^\circ$C,
the household's preferred minimum and maximum, and the current setpoint plus
or minus $0.5^\circ$C. Values are clipped to $[22,28]^\circ$C. During an
active VPP event, candidates below
$\max\{T^{\mathrm{pref,max}}+0.5,T_k^{c,\mathrm{sp}}\}$ are removed when at
least one raised-setpoint candidate remains. \\

Washer, dishwasher, and dryer &
For task $j$, candidates include its preferred time, earliest feasible time,
latest start time, the time immediately before a VPP event, and the event end.
Each candidate must be in the configured service window, must not start in the
past, and must finish before its deadline. A task already running, completed,
or skipped is locked and is not rescheduled. \\

EV &
Candidate charge starts are
$\{0,2,4,6,20,22\}$ hours, with the configured departure time as the charge
end. The forward SOC model penalizes schedules that do not reach the required
departure SOC. \\

EWH &
Candidates include the configured preheat window and several earlier preheat
windows; the selected target is $65^\circ$C. Candidate windows are checked
against the required bath time and service-readiness constraints. \\
\bottomrule
\end{tabular}
\end{table*}

\paragraph{Finite-horizon objective.}
For each candidate, the regional dynamic model generates six predicted states.
The accumulated objective is
\begin{equation}
a_k^\star
=\arg\min_{a_k\in\mathcal A_k}
\sum_{h=1}^{H}J_{k+h}(s_{k+h},a_k),
\end{equation}
where
\begin{equation}
J_k=\alpha_C C_k^{\mathrm{home}}
+\alpha_U D_k^{\mathrm{user}}
+\alpha_G D_k^{\mathrm{grid}}
+\lambda_{\mathrm{slack}}D_k^{\mathrm{slack}}.
\label{eq:mpc_objective}
\end{equation}
The weights are
\begin{align}
(\alpha_C,\alpha_U,\alpha_G)&=
\begin{cases}
(0.30,0.45,0.25), & \text{ordinary step},\\
(0.25,0.40,0.35), & \text{active VPP step},
\end{cases}
\\
\lambda_{\mathrm{slack}}&=100.
\end{align}

The normalized home-energy cost is
\begin{align}
C_k^{\mathrm{home}}
&=\frac{p_k\Delta t
\left(P_k^{\mathrm{HVAC}}+P_k^{\mathrm{task}}
+P_k^{\mathrm{EWH}}+P_k^{\mathrm{EV}}\right)}
{\max\{1,10p_k\Delta t\}}.
\end{align}
The 10-kWh-equivalent denominator prevents the monetary term from dominating
solely because of the tariff scale.

User discomfort is the mean of the active temperature, setpoint-preference,
task-timing, and EV-service components:
\begin{equation}
D_k^{\mathrm{user}}
=\frac{1}{|\mathcal K_k|}
\sum_{\ell\in\mathcal K_k}d_{\ell,k}.
\end{equation}
The occupied temperature and setpoint terms are
\begin{align}
d_{T,k}
={}&\frac{o_k\Delta t\left(
[T^{\min}-T_k^a]_+^2+[T_k^a-T^{\max}]_+^2\right)}
{\max\{1,(T^{\max}-T^{\min})^2\}},\\
d_{\mathrm{sp},k}
={}&o_k\Delta t
\left[
\frac{|T_k^{c,\mathrm{sp}}-T^{\mathrm{pref}}|}
{\Delta T^{\mathrm{acc}}}-1
\right]_+^2,
\end{align}
where $o_k$ is the occupancy probability and
$\Delta T^{\mathrm{acc}}=1^\circ$C unless overridden by the household
configuration. For a shiftable task $j$,
\begin{equation}
d_{j,k}=w_j
\left(
\frac{[s_j^{\mathrm{pref}}-s_j]_+
+[f_j-f_j^{\mathrm{pref}}]_+}
{W_j}
\right)^{q},
\qquad q=2,
\end{equation}
where $W_j$ is its allowed service-window length. Explicitly skipping a
required task incurs unit timing discomfort in addition to service slack. The
EV discomfort is the squared normalized departure-SOC shortfall.

The grid term penalizes predicted power above the available event or grid
target:
\begin{equation}
D_k^{\mathrm{grid}}
=
\frac{[P_k^{\mathrm{total}}-P_k^{\mathrm{target}}]_+^2
\Delta t}
{\max\{\epsilon,(P_k^{\mathrm{target}})^2\}}.
\end{equation}
The implementation retains a separate realized-DR contribution field, but
sets it to zero at planning time because the realized reduction and post-event
rebound are not yet observable. Thus, no post-event information enters the
ex-ante MPC decision.

\paragraph{Constraints and slack.}
Action enumeration enforces device presence, setpoint limits, task service
windows, task duration, scheduling after the current time, EWH timing, and
EV charging windows. Remaining violations are represented by squared slack
terms:
\begin{equation}
\begin{aligned}
D_k^{\mathrm{slack}}
={}&D_k^{\mathrm{HVAC-bound}}
+D_k^{\mathrm{temperature-safety}}\\
&+D_k^{\mathrm{task-skip/deadline}}
+D_k^{\mathrm{EWH-service}}\\
&+D_k^{\mathrm{EV-SOC}}
+D_k^{\mathrm{grid-limit}}.
\end{aligned}
\end{equation}
A slack component is activated only when the corresponding physical field is
available. The MPC objective excludes simulated acceptance, role-play scores,
language-model cost, latency, token consumption, and final benchmark metrics.

\subsection{Rule+MILP}
\label{sec:rule_milp_details}

Rule+MILP separates HVAC selection from appliance scheduling. HVAC is selected
by regional dynamic-model enumeration, while non-HVAC services are scheduled
by a binary mixed-integer program on a 30-minute grid. This design makes the
service constraints explicit and avoids introducing language-derived
preferences into the physical controller.

\paragraph{HVAC rule and objective.}
The HVAC module uses the same regional 5R3C model and six-step, one-hour
prediction horizon as MPC Dynamic. In the standalone baseline, ordinary
cooling-setpoint candidates are
\begin{equation}
\mathcal T=\{22.0,22.5,\ldots,28.0\}\ ^\circ\mathrm{C}.
\end{equation}
When the prediction horizon overlaps a VPP event, a
$40^\circ$C cooling-setpoint sentinel is additionally available to represent
HVAC-off control. For candidate setpoint $T$, the standalone HVAC objective is
\begin{equation}
J^{\mathrm{HVAC}}(T)
=p_k E^{\mathrm{HVAC}}(T)
+10^{4}E^{\mathrm{HVAC,VPP}}(T),
\label{eq:rule_hvac_objective}
\end{equation}
where the two energy quantities are obtained from the six-step dynamic
rollout. Consequently, the controller strongly prioritizes suppressing HVAC
operation inside the event window. Temperature and PMV remain evaluation
outcomes rather than terms in the standalone Rule+MILP objective. If the
regional model cannot be loaded, the implementation falls back to a PMV rule
and selects the warmest candidate satisfying $|\mathrm{PMV}|\leq0.5$, when
such a candidate exists.

\paragraph{MILP variables and objective.}
Let $\mathcal I$ be the set of present, unlocked appliances and
$\mathcal C_i$ the feasible candidates generated for appliance $i$. For every
candidate $c\in\mathcal C_i$, define
\begin{equation}
x_{ic}\in\{0,1\},
\end{equation}
where $x_{ic}=1$ means that candidate $c$ is selected. Its energy cost is
computed on the 30-minute tariff grid:
\begin{equation}
C_{ic}
=\sum_{\tau\in\mathcal H_{ic}}
p_\tau P_{ic}\Delta\tau.
\end{equation}
The candidate objective is
\begin{equation}
\widetilde C_{ic}
=C_{ic}
+10^{4}\,
\mathbb I\!\left[
\mathcal H_{ic}\cap\mathcal H^{\mathrm{VPP}}\neq\varnothing
\right].
\end{equation}
The scheduling problem is
\begin{align}
\min_{\{x_{ic}\}}\quad&
\sum_{i\in\mathcal I}\sum_{c\in\mathcal C_i}
\widetilde C_{ic}x_{ic},\\
\mathrm{s.t.}\quad&
\sum_{c\in\mathcal C_i}x_{ic}=1,
\qquad \forall i\in\mathcal I,\\
&x_{ic}\in\{0,1\}.
\label{eq:rule_milp}
\end{align}
The current problem contains one independent choose-one constraint per
appliance and therefore decomposes by appliance. It is nevertheless solved
through the same binary MILP interface using CBC. If the MILP package is
unavailable, exact enumeration returns the same minimum. Equal-objective
candidates are sampled uniformly so that the baseline does not silently add an
unmodeled preferred start time; this affects only exact ties.

\paragraph{Candidate generation and constraints.}
For washer, dishwasher, and dryer, candidate starts are placed at 30-minute
intervals satisfying
\begin{align}
s_i&\geq \max\{t_k,e_i\},\\
s_i+d_i&\leq l_i,\\
s_i+d_i&<t^{\mathrm{run-end}},
\end{align}
where $e_i$, $l_i$, and $d_i$ are the earliest time, latest finish, and
duration. Overnight windows are converted to absolute simulation time before
these tests. To ensure that completion is recorded before the simulator stops,
one additional 30-minute grid interval is retained before a finite run
boundary. Non-shiftable services use their configured fixed start.

For the EWH, preheat candidates extend from the current time to the required
bath deadline. The preheat duration is taken from the configured service
window, the target temperature is $63^\circ$C, and the energy-cost proxy scales
rated power by
\begin{equation}
\frac{63-40}{65-40}.
\end{equation}
For the EV, the required charging energy is
\begin{equation}
E^{\mathrm{EV,need}}
=\max\left\{
E^{\mathrm{daily-drive}},
[z^{\mathrm{target}}-z_k^{\mathrm{EV}}]_+
C^{\mathrm{EV}}
\right\},
\end{equation}
and the charging duration is rounded upward to the nearest 30 minutes:
\begin{equation}
d^{\mathrm{EV}}
=0.5\left\lceil
\frac{E^{\mathrm{EV,need}}}
{0.5\,\eta^{\mathrm{EV}}\bar P^{\mathrm{EV}}}
\right\rceil.
\end{equation}
EV candidates must fit before the daily and simulation cutoffs.

Whenever at least one candidate for an appliance avoids every VPP window, all
VPP-overlapping candidates for that appliance are removed before optimization.
The $10^4$ penalty is retained for diagnostics and for the case in which no
VPP-safe service schedule exists. An appliance is locked for the remainder of
the day once it is running, completed, skipped, ready at the required bath
time, or has reached its EV target.

The complete Rule+MILP objective reported in the controller trace is
\begin{equation}
J^{\mathrm{Rule+MILP}}
=J^{\mathrm{HVAC}}
+\sum_{i,c}\widetilde C_{ic}x_{ic}.
\end{equation}
No simulated acceptance or post-event capacity result is used during this
optimization.

\subsection{PPO Controller}
\label{sec:ppo_details}

\paragraph{Markov decision process.}
The PPO baseline is formulated as a finite-horizon Markov decision process
$\langle\mathcal S,\mathcal A,\mathcal P,r,\gamma\rangle$ with a 10-minute
decision interval and discount factor $\gamma=0.995$. The policy maximizes
\begin{equation}
\max_\theta\;
\mathbb E_{\pi_\theta}
\left[\sum_{k=0}^{K-1}\gamma^k r_k+\gamma^K R_{\mathrm{terminal}}\right].
\end{equation}
Regional PPO checkpoints are trained in the fast dynamic environment using a
single-step transition of the same regional 5R3C model. Final benchmark
evaluation uses deterministic policy inference in the common EnergyPlus
environment.

\paragraph{Observation space.}
The observation is a 41-dimensional real vector. Its exact composition is
given in Table~\ref{tab:ppo_observation}.

\begin{table*}[t]
\centering
\caption{PPO observation vector. Scaling is applied before the vector is passed
to the policy network.}
\label{tab:ppo_observation}
\small
\setlength{\tabcolsep}{4pt}
\renewcommand{\arraystretch}{1.12}
\begin{tabular}{p{0.16\textwidth}p{0.08\textwidth}p{0.66\textwidth}}
\toprule
Group & Dim. & Features and scaling \\
\midrule
Time &
4 &
$\sin(2\pi h/24)$, $\cos(2\pi h/24)$, day index divided by 7, and time to the
next 18:00 VPP event divided by 24. \\

Thermal and occupancy &
4 &
Indoor temperature divided by 40, outdoor temperature divided by 45, current
setpoint divided by 30, and the binary occupancy indicator
$\mathbb I[8{:}00\leq h<22{:}00]$. \\

VPP context &
4 &
Active-event indicator, scaled event target, committable capacity divided by
2, and recommended bid divided by 2. Capacity values are set to zero outside
an active event. \\

Price &
5 &
Current price and the next-six-hour mean, maximum, and minimum prices, each
divided by 0.3, together with a peak-price indicator. The look-ahead statistics
use 30-minute samples over the next six hours. \\

Preference proxy &
4 &
Comfort weight, price sensitivity, task flexibility, and VPP cooperation,
derived from the structured persona tags and scoring weights. \\

Washer &
5 &
Presence, execution state divided by 3, scheduled hour divided by 24, earliest
hour divided by 24, and latest hour divided by 24. Execution state encodes
absent/skipped, waiting, running, and completed states. \\

Dishwasher &
5 &
The same five fields as the washer. \\

EWH &
5 &
Presence, preheat-request indicator, preheat start and end divided by 24, and
required bath time divided by 24. \\

EV &
3 &
Presence, current SOC, and at-home indicator. \\

Refrigerator &
2 &
Presence and power divided by 2. The refrigerator is observed but is not
directly controlled. \\
\midrule
Total & 41 & The dryer has an action coordinate but no separate observation
block; its valid interval and execution constraints are enforced by the action
decoder and daily scheduling adapter. \\
\bottomrule
\end{tabular}
\end{table*}

\paragraph{Action space.}
The neural policy outputs
\begin{equation}
u_k\in[-1,1]^8.
\end{equation}
Each coordinate is linearly decoded into the physical action shown in
Table~\ref{tab:ppo_action}.

\begin{table*}[t]
\centering
\caption{PPO action vector and physical decoding ranges.}
\label{tab:ppo_action}
\small
\setlength{\tabcolsep}{4pt}
\renewcommand{\arraystretch}{1.10}
\begin{tabular}{cllc}
\toprule
Dim. & Physical action & Decoder & Range \\
\midrule
0 & HVAC cooling setpoint & $25+3u_0$ & $[22,28]^\circ$C \\
1 & Washer start hour & $13.5+5.5u_1$ & $[8,19]$ h \\
2 & Dishwasher start hour & $20.25+1.25u_2$ & $[19,21.5]$ h \\
3 & EWH preheat start & $12+5u_3$ & $[7,17]$ h \\
4 & EWH target temperature & $60+15u_4$ & $[45,75]^\circ$C \\
5 & EV charge start & $19.25+0.75u_5$ & $[18.5,20]$ h \\
6 & EV charge end & $5.75+1.75u_6$ & $[4,7.5]$ h \\
7 & Dryer start hour & $13.75+5.75u_7$ & $[8,19.5]$ h \\
\bottomrule
\end{tabular}
\end{table*}

The HVAC setpoint is applied at every decision. Appliance schedules are
accepted at most once per device per day. Outputs for absent devices are
ignored. For washer, dishwasher, and dryer, a non-DR-adjustable household
configuration overrides the policy output with the household's preferred
routine time. Their scheduling commands are withheld when the decision itself
occurs inside an active VPP window. EWH preheat lasts three hours, with its end
capped at 18:00. EV mode is fixed to \texttt{smart}; its start is additionally
clipped to the household arrival time and its end to the allowed pre-departure
range.

\paragraph{Step reward.}
Let $e_k=P_k^{\mathrm{total}}\Delta t$ be step energy, $p_k$ the current price,
$s^{\mathrm{price}}\in[0,1]$ the price-sensitivity proxy, and
$s^{\mathrm{VPP}}\in[0,1]$ the VPP-cooperation proxy. The occupied comfort
violation is
\begin{equation}
d_k^T=
[T^{\min}-T_k^a]_+
+[T_k^a-T^{\max}]_+,
\end{equation}
where the comfort band is read from the household configuration. The
price multiplier is
\begin{equation}
m_k^p
=1+0.2s^{\mathrm{price}}
\max\left(0,\frac{p_k}{0.15}\right).
\end{equation}
The step reward used by the released checkpoints is
\begin{equation}
\begin{aligned}
r_k
={}&
\operatorname{clip}\Bigl(
-\bigl[
0.3e_km_k^p
+8\,o_kd_k^T
+2s^{\mathrm{VPP}}v_ke_k
\bigr],\\
&\qquad -50,\;50
\Bigr),
\end{aligned}
\label{eq:ppo_step_reward}
\end{equation}
where $o_k=\mathbb I[8{:}00\leq h_k<22{:}00]$ and $v_k$ is the active-VPP
indicator. Thus PPO is penalized for energy consumption, price-weighted energy,
occupied comfort violations, and event-window consumption.

\paragraph{Terminal reward.}
The episode-end reward represents service fulfillment and VPP avoidance:
\begin{equation}
\begin{split}
R_{\mathrm{terminal}}={}&
200f_{\mathrm{washer}}
+200f_{\mathrm{dishwasher}}
+200f_{\mathrm{dryer}}\\
&+100f_{\mathrm{EWH}}
+300f_{\mathrm{EV}}
+100f_{\mathrm{avoid}}\\
&+\max\{0,80-3E^{\mathrm{VPP}}\}.
\end{split}
\label{eq:ppo_terminal_reward}
\end{equation}
Here, the three shiftable-load fractions measure schedules issued early enough
to finish before their configured latest times;
$f_{\mathrm{EWH}}$ is the fraction of required bath events for which hot water
is ready; $f_{\mathrm{EV}}$ is the fraction reaching target SOC; and
$f_{\mathrm{avoid}}$ is the fraction of present shiftable and EWH services
that are scheduled feasibly without operating during a VPP event.
$E^{\mathrm{VPP}}$ is total episode energy consumed inside all VPP windows.

\paragraph{PPO optimization and checkpoint settings.}
The policy and value functions use two-layer multilayer perceptrons with
256 hidden units per layer. PPO uses a rollout length of 432 steps per
environment, batch size 2,048, learning rate $3\times10^{-4}$,
$\gamma=0.995$, GAE parameter 0.95, clipping parameter 0.2, ten optimization
epochs per update, entropy coefficient 0, value-loss coefficient 0.5, and
maximum gradient norm 0.5. Thirty-two dynamic-model environments are run in
parallel. Both regional checkpoints were requested for ten million training
steps and contain 10,008,576 steps because the final vectorized rollout crosses
the requested boundary.

The PPO surrogate is
\begin{equation}
\begin{aligned}
L^{\mathrm{clip}}(\theta)
={}&
\mathbb{E}_k\Bigl[
\min\Bigl(
\rho_k(\theta)\widehat{A}_k, \\
&\qquad
\operatorname{clip}\bigl(
\rho_k(\theta),1-\epsilon,1+\epsilon
\bigr)\widehat{A}_k
\Bigr)
\Bigr].
\end{aligned}
\end{equation}
with $\epsilon=0.2$ and
$\rho_k(\theta)=
\pi_\theta(a_k\mid s_k)/\pi_{\theta_{\mathrm{old}}}(a_k\mid s_k)$.
The saved training configuration uses advantage normalization and ten epochs
per rollout.

\section{Ordinary-Routine Physical Reference}

We additionally execute the same five households and two regions for seven
days under an ordinary no-DR routine. This physical reference quantifies the
event-window load retained without a VPP-specific household plan and is reported
separately from the agent/controller rankings.

\begin{table}[H]
\centering

\resizebox{\columnwidth}{!}{
\begin{tabular}{lrrrrr}
\toprule
Region & No-DR kWh/h & EB kWh/h & Reduction & No-DR cost & EB cost \\
\midrule
Tianjin & 5.489 & 1.597 & \textbf{70.9\%} & 63.242 & 54.681 \\
Berlin & 5.503 & 1.605 & \textbf{70.8\%} & 4.432 & 3.842 \\
\bottomrule
\end{tabular}}

\caption{EnergyBridge against the separately executed ordinary-routine
reference. Cost uses each region's native unit.}
\label{tab:nodr_reference}
\end{table}

EnergyBridge removes about 71\% of ordinary-routine event-window energy in
both regions. Daily cost also
falls by 13.5\% in Tianjin and 13.3\% in Berlin. This separate reference makes
the physical meaning of the 1.60~kWh/h main result directly observable without
mixing a non-participating routine into the consent ranking.

\section{Agent Interaction Resource Accounting}

The frozen logs record every language-model call and token. We report measured
resource counts rather than converting them through a model-price assumption,
which keeps the comparison independent of changing API prices.

\begin{table}[H]
\centering

\resizebox{\columnwidth}{!}{
\begin{tabular}{llrrrr}
\toprule
Region & Method & Calls & Prompt tok. & Completion tok. & Total tok. \\
\midrule
Tianjin & EnergyBridge & 7.0 & 108,451 & 5,252 & 113,703 \\
 & HEMA & 4.0 & 98,307 & 2,627 & 100,934 \\
Berlin & EnergyBridge & 7.0 & 108,961 & 4,897 & 113,858 \\
 & HEMA & 4.0 & 96,152 & 2,666 & 98,818 \\
\bottomrule
\end{tabular}}
\caption{Measured LLM resources per household-day. Both agent methods use the
same 35 household-days per region.}
\label{tab:llm_resources}
\end{table}

The additional interaction budget corresponds to onboarding, skill selection,
event-level proposal generation, and feedback-memory updates. Alongside
12.7-15.2\% more tokens than HEMA, EnergyBridge shows observed regional
differences of 0.786-0.854 in score, 40.0-42.9 percentage points in simulated
gate acceptance, 23.4-35.2\% in VPP-window energy, and 5.1-9.7\% in
electricity cost. Together, these values make the observed
performance-interaction trade-off explicit.

\section{Capacity Retrieval Validation and Sensitivity}

We evaluate the June-to-July temporal capacity validation for
$k\in\{1,3,5,7,10,15\}$. Every row evaluates the same 208 accepted events; only
the number of historical neighbors changes.

\begin{table}[H]
\centering

\resizebox{\columnwidth}{!}{
\begin{tabular}{rrrrrr}
\toprule
$k$ & $n$ & Retrieval pass & MAE (kWh) & Mean ratio & IS pass \\
\midrule
1  & 208 & 85.10\% & \textbf{0.561} & 0.980 & 85.10\% \\
3  & 208 & 86.06\% & 0.574 & 0.971 & 86.06\% \\
5  & 208 & \textbf{86.54\%} & 0.578 & 0.946 & \textbf{86.54\%} \\
7  & 208 & 84.13\% & 0.590 & 0.939 & 84.13\% \\
10 & 208 & 81.25\% & 0.600 & 0.924 & 81.25\% \\
15 & 208 & 77.40\% & 0.621 & 0.916 & 78.37\% \\
\bottomrule
\end{tabular}
}
\caption{Capacity sensitivity to retrieval size on the July validation cohort.
Pass denotes $0.8\leq C^{actual}/\widehat C\leq1.2$; MAE is in kWh.}
\label{tab:capacity_topk}
\end{table}

Within this validation sensitivity set, top-5 gives the best target-band coverage, while
top-1 minimizes MAE by using the closest historical analogue. EnergyBridge
reports top-5 because the benchmark prioritizes band coverage rather than only
average absolute error. Compact neighborhoods remain stable under importance
weighting, which adds 0.97 points for the wider $k=15$ neighborhood.

\end{document}